\documentclass[conference]{IEEEtran}
\IEEEoverridecommandlockouts
\usepackage{cite}
\usepackage{blindtext}
\usepackage{graphicx}
\usepackage{amsfonts}
\usepackage{color}
\usepackage{xcolor}
\usepackage{amsmath}
\usepackage{stmaryrd}
\usepackage{algorithm}
\usepackage{algpseudocode}
\usepackage{hyperref}
\usepackage{amsthm} 
\usepackage{mathtools}
\usepackage{bm}
\usepackage{booktabs}
\usepackage{colortbl} 
\usepackage{placeins}
\usepackage{lipsum}

\def\BibTeX{{\rm B\kern-.05em{\sc i\kern-.025em b}\kern-.08em
    T\kern-.1667em\lower.7ex\hbox{E}\kern-.125emX}}

\DeclareMathAlphabet\ten{OMS}{cmsy}{b}{n}
\DeclareMathAlphabet\mathcalbf{OMS}{cmsy}{b}{n}

\newtheorem{remark}{Remark}

\begin{document}

\title{Tensor-Train Recurrent Neural Networks for Interpretable Multi-Way Financial Forecasting}


\author{\IEEEauthorblockN{Yao Lei Xu, Giuseppe G. Calvi, Danilo P. Mandic}
\IEEEauthorblockA{\textit{Department of Electrical and Electronic Engineering} \\
\textit{Imperial College London}\\
London, United Kingdom \\
\{yao.xu15, giuseppe.calvi15,  d.mandic\}@imperial.ac.uk}
}

\maketitle

\begin{abstract}
Recurrent Neural Networks (RNNs) represent the \textit{de facto} standard machine learning tool for sequence modelling, owing to their expressive power and memory. However, when dealing with large dimensional data, the corresponding exponential increase in the number of parameters imposes a computational bottleneck. The necessity to equip RNNs with the ability to deal with the curse of dimensionality, such as through the parameter compression ability inherent to tensors, has led to the development of the Tensor-Train RNN (TT-RNN). Despite achieving promising results in many applications, the full potential of the TT-RNN is yet to be explored in the context of interpretable financial modelling, a notoriously challenging task characterized by multi-modal data with low signal-to-noise ratio. To address this issue, we investigate the potential of TT-RNN in the task of financial forecasting of currencies. We show, through the analysis of TT-factors, that the physical meaning underlying tensor decomposition, enables the TT-RNN model to aid the interpretability of results, thus mitigating the notorious ``black-box" issue associated with neural networks. Furthermore, simulation results highlight the regularization power of TT decomposition, demonstrating the superior performance of TT-RNN over its uncompressed RNN counterpart and other tensor forecasting methods.
\end{abstract}


\section{Introduction}

Recurrent Neural Networks (RNNs) \cite{Mandic2001} have proven to be among the most successful machine learning approaches for sequence modelling. Despite considerable success, RNNs are known to be notoriously difficult to train when considering large-dimensional inputs, due to an exponential increase in the number of parameters \cite{He2017, khrulkov2017expressive}. One way to tackle this issue is through the compression properties of tensor decomposition.

Tensors are a multi-linear generalization of vectors and matrices to multi-way arrays \cite{Mandic2015, Kolda2009}. Similar to their matrix counterparts, latent factors in tensors can be extracted via tensor decomposition (TD), which can be employed for compression and to obtain low-rank representations of multi-way data \cite{Kolda2009, sidiropoulos2017tensor}. These desirable properties, combined with the fact that tensors can be readily constructed from vectors and matrices via \textit{tensorization} \cite{Mandic2016_2, Mandic2017}, make them particularly attractive to the data analytics communities. 

Although substantial research has been carried out on the topic \cite{Kim2015, Zhong2019, Calvi2019_2, lebedev2014speeding}, tensors have firstly been considered in conjunction with neural networks only in 2015 \cite{Novikov2015}, where the authors employed Tensor-Train Decomposition (TTD) \cite{Oseledets2011} to drastically compress a fully-connected neural network, while maintaining comparable performance. This was later extended to TT-RNNs in \cite{Yang2017, yu2017, tjandra2017compressing} for modelling sequential data. Despite promising results, the full potential stemming from the combination of tensors and neural networks is yet to be explored, especially in the area financial forecasting.

Financial forecasting is a notoriously difficult task, characterized by multi-way data with low signal-to-noise ratio \cite{de2020machine, de2018advances, dees2019analysing}. In the high-frequency trading domain, several deep learning and tensor based algorithms have demonstrated promising results by learning from large amounts of order-book data \cite{tsantekidis2017using, tran2018temporal, tran2017tensor, dixon2018sequence, tsantekidis2020using}. However, at a daily trading frequency, the amount of available data becomes too scarce compared to its dimensionality, which poses significant challenges to these learning algorithms in terms of over-fitting \cite{gal2015theoretically, dieng2018noisin}. 

We here show that the TT-RNN offers a plausible solution to the notoriously difficult task of low-frequency financial forecasting. Indeed, the multi-way nature of financial data naturally admits a tensor representation, which is inherent to the tensorized nature of TT-RNN. In addition, the super-compression property of the TTD drastically reduces the parameter complexity of the RNN, which can be used as a regularization tool that directly controls over-fitting, while preserving the underlying data structure and interpretability. This motivates us to employ a TT-RNN to the task of financial forecasting of currencies, a task which involves multi-way financial data from a range of assets and asset classes. We also show how the inherent expressiveness of tensors enables the TT-RNN to offer new perspectives on the interpretability of results, by indicating which TT-cores, and consequently which tensor modes \cite{Calvi2019_2}, play a more prominent role in forecasting. This both mitigates the ``black-box" nature of neural networks, and offers enhanced results over its uncompressed RNN counterpart and other tensor models, as illustrated in experiments.

\section{Theoretical Background}

A short background necessary for this work is given below, and is based on the work in \cite{He2017,Novikov2015, Yang2017}. The tensor indices in this paper are grouped according to the Little-Endian convention \cite{Dolgov2014}.

\subsection{Tensor Notation and Operations}

\begin{table}[h]
	\centering
    \caption{Tensor and matrix nomenclature}
    \begin{tabular}{ll}
    	\hline \vspace{-1mm}\\
    	
    	$\mathcalbf{X} \in \mathbb{R}^{I_1 \times I_2 \times \cdots \times I_N}$ & \begin{tabular}[c]{@{}l@{}} $N$-th order tensor of size \\ $I_1 \times I_2 \times \cdots \times I_N$\end{tabular} \vspace{2mm} \\
    	
        $\mathbf{X}  \in \mathbb{R}^{I_1 \times I_2}$ & Matrix of size $I_1 \times I_2$  \vspace{2mm}\\
    	
    	$\mathbf{x} \in \mathbb{R}^{I_1}$ & Vector of size $I_1$  \vspace{2mm}\\
    	
    	$x \in \mathbb{R}$ & Scalar \vspace{2mm}\\
    	
    	$x_{i_1, i_2, \cdots, i_N}$ 	& $(i_1,i_2,\ldots,i_N)$ entry of $\mathcalbf{X}$ \vspace{2mm} \\
    	
        \hline
    \end{tabular}
	\label{table:nomenclature}
\end{table}

An order-$N$ tensor, $\ten{X}$ $\in \mathbb{R}^{I_1 \times I_2 \times \cdots \times I_N}$, is an $N$-way array with $N$ modes, where mode-$n$ is of dimensionality $I_n$, $n=1, 2, \dots, N$. Special instances of tensors include matrices ($\mathbf{X}  \in \mathbb{R}^{I_1 \times I_2}$) as order-$2$ tensors, vectors ($\mathbf{x} \in \mathbb{R}^{I_1}$) as order-$1$ tensors, and scalars ($x \in \mathbb{R}$) as order-$0$ tensors. The $(i_1,i_2,\ldots,i_N)$ scalar entry of a tensor, $\ten{X}(i_1, i_2, \dots, i_N)$, is denoted as $x_{i_1,i_2,\cdots,i_N}$. A tensor can be unfolded into a matrix via the \textit{matricization} process, while the reverse process is referred to as \textit{tensorization}. 

\begin{figure}[h]
	\centering
	\includegraphics[width=1\columnwidth]{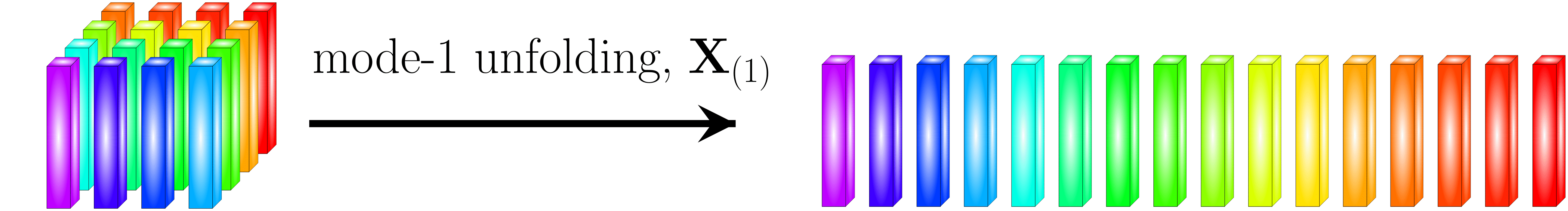}
\end{figure}%
\begin{figure}[h]
	\centering
	\vspace{-5mm}
	\includegraphics[width=1\columnwidth]{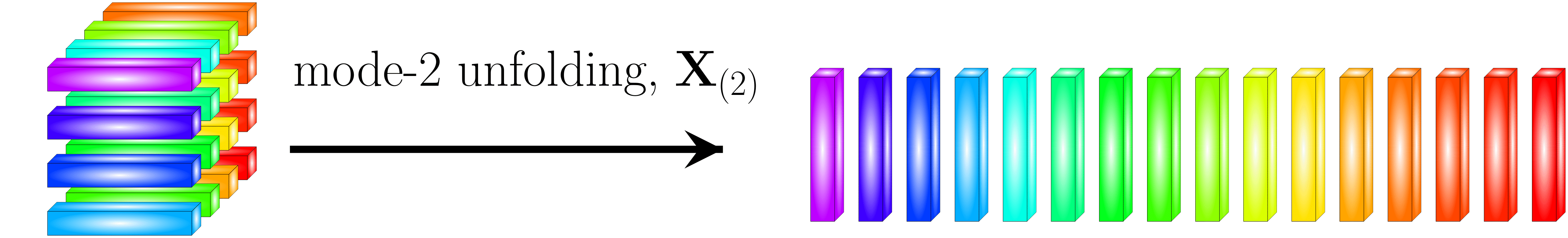}
\end{figure}
\begin{figure}[h]
	\centering
	\vspace{-5mm}
	\includegraphics[width=1\columnwidth]{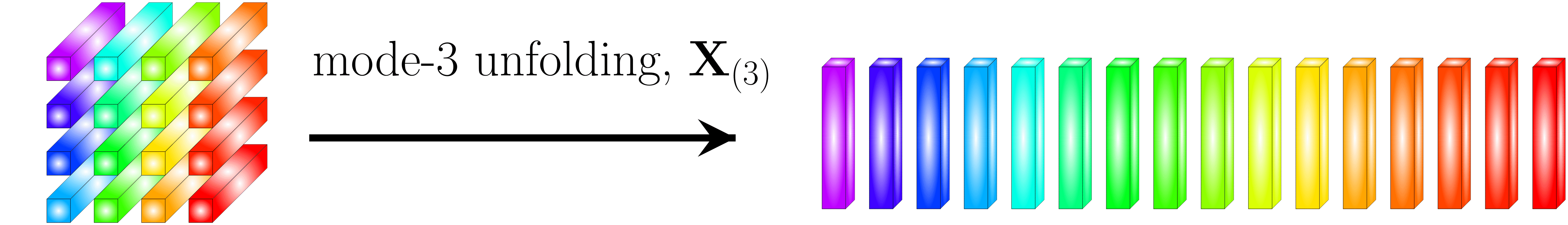}
	\caption{matricization of a $3$-rd order tensor.}
	\label{fig:unfolding}
\end{figure}

An $(m,n)$ tensor contraction \cite{Kolda2009, Cichocki2014} between an $N$-th order tensor, $\ten{A} \in \mathbb{R}^{I_1\times \cdots \times I_n \times \cdots \times I_N}$, and an $M$-th order tensor, $\ten{B}\in \mathbb{R}^{J_1\times \dots \times J_m \times \dots \times J_M} $, with $I_n = J_m$, is denoted by $\times^m_n$, which yields an $(N+M-2)$-th order tensor, $\ten{C}\in \mathbb{R}^{I_1 \times \cdots \times I_{n-1} \times I_{n+1}  \times \cdots \times I_N \times J_1 \times \cdots \times J_{m-1} \times J_{m+1}  \times \cdots \times J_M}$, with entries
\begin{equation}\label{eq:cont}
	\begin{aligned}
		&c_{i_1,\dots,i_{n-1}, i_{n+1}, \dots, i_N, j_1, \dots, j_{m-1}, j_{m+1}, \dots, j_M   } \\
		&= \sum_{i_n=1}^{I_n} a_{i_1, \dots, i_{n-1}, i_n, i_{n+1}, \dots, i_N } b_{j_1, \dots, j_{m-1}, i_n, j_{m+1}, \dots, j_M}   
	\end{aligned}
\end{equation}

\subsection{Tensor-Train Decomposition}

\begin{figure}[t!]
	\centering
	\includegraphics[width=1\linewidth]{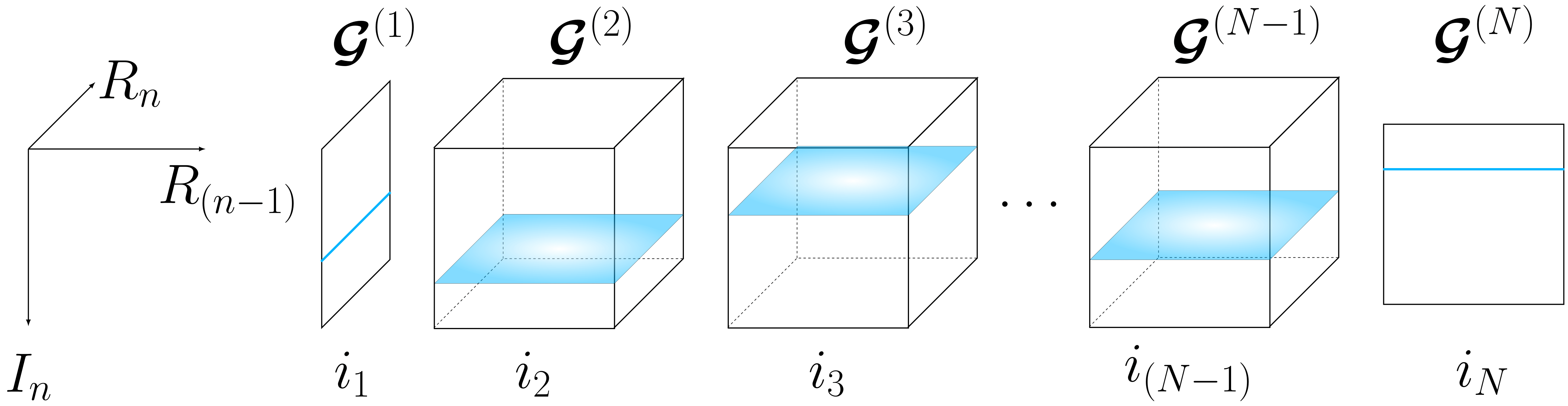}
	\caption{$N$-th order TT-decomposition according to (\ref{eq:tt}).}
	\label{fig:tt1}
\end{figure}

\begin{figure}[t!]
	\centering
	\includegraphics[width=1\linewidth]{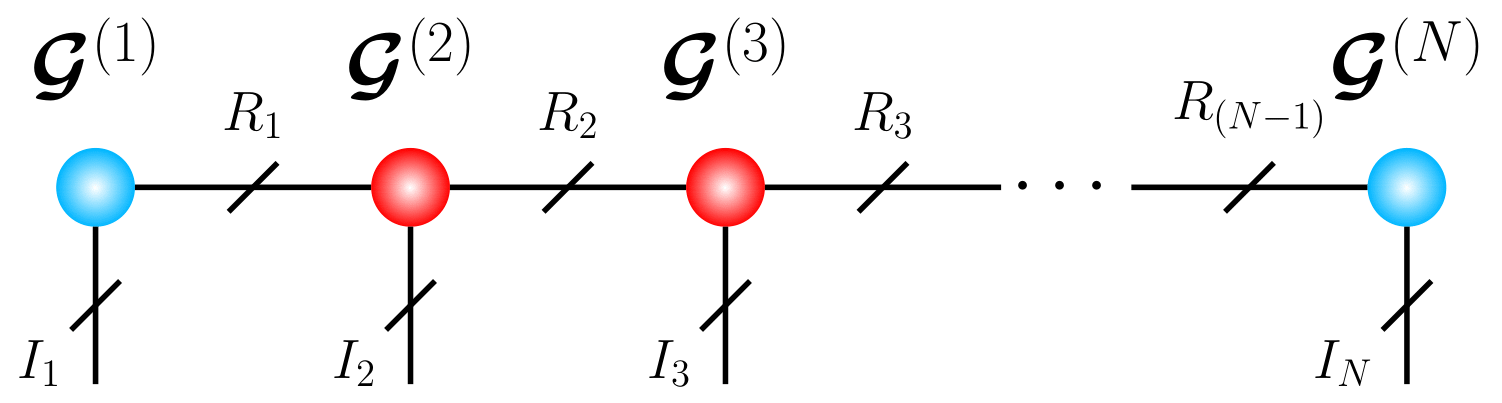}
	\caption{$N$-th order TT-decomposition according to (\ref{eq:ttmult}).}
	\label{fig:tt2}
	\vspace{-5mm}
\end{figure}

Tensor Train (TT) decomposition \cite{Oseledets2009} \cite{Oseledets2011} was introduced by Oseledets to help mitigate the computational bottlenecks that arise from the curse of dimensionality, as tensor algorithms can become intractable for high order tensors. The most common form of TT is the Matrix Product State (MPS or TT-MPS), introduced in the quantum physics community \cite{Cichocki2014}, which decomposes a large $N$-th order tensor, $\ten{X} \in \mathbb{R}^{K_1 \times K_2 \times \cdots \times K_N}$, into $N$ smaller $2$-nd or $3$-rd order core tensors $\ten{G}^{(n)}$, given by
\begin{equation}
\ten{G}^{(n)} \in \mathbb{R}^{ R_{n-1} \times  K_n \times R_n }, \hspace{2mm} n=1, \dots, N
\end{equation}
The tensors $\ten{G}^{(1)}, \ldots, \ten{G}^{(N)}$ are referred to as \textit{core tensors}, while the set $(R_0, \dots, R_{N})$, where $R_0=R_N=1$, represents the \textit{TT-rank} of the TT decomposition.

By defining $\ten{G}^{(n)}(k_n)$, $k_n = 1, \dots, K_N$ as the $k_n$-th horizontal slice of tensor $\ten{G}_n$, the MPS assumes the element-wise form in (\ref{eq:tt}). The corresponding slice-wise matrix multiplication in (\ref{eq:tt}) is illustrated in Figure \ref{fig:tt1}.
\begin{equation} \label{eq:tt}
\begin{aligned}
\ten{X}(k_1, k_2, \dots, k_N) &= \ten{G}^{(1)}(k_1)\ten{G}^{(2)}(k_2)\cdots\ten{G}^{(N)}(k_N)
\end{aligned}
\end{equation}

The TT decomposition in (\ref{eq:tt}) may also be expressed as a series of tensor contractions as in (\ref{eq:ttmult}). Figure \ref{fig:tt2} illustrates the Tensor-Network representation \cite{Cichocki2014} of (\ref{eq:ttmult}), where tensors are represented as nodes of a graph, while the edges represent a linear contraction over modes of common dimensionality.

\begin{equation}\label{eq:ttmult}
	\begin{aligned}
		\ten{X} &= \ten{G}^{(1)} \times^1_2 \ten{G}^{(2)} \times^1_3 \ten{G}^{(3)} \times^1_3 \cdots \times^1_3 \ten{G}^{(N)}\\
		&=\llparenthesis \ten{G}^{(1)}, \ten{G}^{(2)}, \dots, \ten{G}^{(N)}   \rrparenthesis
	\end{aligned}
\end{equation}

\subsection{Tensor-Train Fully Connected Neural Networks}

A fully connected layer in neural networks applies an element-wise activation function, $\sigma ( \cdot )$, on the linear transformation to an input vector $\mathbf{x} \in \mathbb{R}^{P}$, that is
\begin{equation}\label{eq:lin}
	\mathbf{y} = \mathbf{W}\mathbf{x} + \mathbf{b}
\end{equation}
with a weight matrix, $\mathbf{W} \in \mathbb{R}^{M \times P}$, and a bias vector, $\mathbf{b} \in \mathbb{R}^M$. If the dimensions of the typically large weight matrix can be factorized as $P = \prod_{n=1}^{N} I_n$ and $M=\prod_{n=1}^{N}J_n$, for $i_n = 1, \dots, I_n$ and $j_n = 1, \dots, J_n$, then the linear transformation (\ref{eq:lin}) can be expressed in tensor form as: 
\begin{equation}\label{eq:ttrep}
	\begin{aligned}
		&\ten{Y}(j_1, \dots, j_N) =\\
		 &=\sum_{i_1, \dots, i_N} \ten{W} ((i_1,j_1), \ldots, (i_N,j_N)) \ten{X}(i_1,i_2,\ldots,i_N)  \\
		 & \hspace{5mm}+ \ten{B}(j_1, \dots, j_N)
	\end{aligned}
\end{equation}
where $\ten{X} \in \mathbb{R}^{I_1  \times \cdots \times I_N}$, $\ten{Y} \in \mathbb{R}^{J_1  \times \cdots \times J_N}$, $\ten{B} \in \mathbb{R}^{J_1  \times \cdots \times J_N}$ are order-$N$ tensorizations of $\mathbf{x} \in \mathbb{R}^P, \mathbf{y} \in \mathbb{R}^M$, and $\mathbf{b} \in \mathbb{R}^M$ respectively, while $\ten{W} \in \mathbb{R}^{I_1 \times J_1 \times \cdots \times I_N \times J_N}$ is the order-$2N$ tensorization of $\textbf{W} \in \mathbb{R}^{M \times P}$ \cite{Yang2017}.

The so tensorized weight matrix can now be expressed in the \textit{TT-format} through a multi-indexing scheme, which extends the MPS form in (\ref{eq:tt}) to the Matrix-Product-Operator (MPO) form in (\ref{eq:ttformat}), where $\ten{G}_n$ are effectively $3$-rd and $4$-th order core tensors (see \cite{Novikov2015} \cite{Yang2017} for more details).
\begin{equation}\label{eq:ttformat}
	\begin{aligned}
		\ten{W} ((i_1,j_1), \ldots, (i_N,j_N)) = \ten{G}^{(1)}(i_1, j_1) \cdots \ten{G}^{(N)}(i_N, j_N) 
	\end{aligned}
\end{equation} 

\subsection{Tensor-Train Recurrent Neural Networks}

Recurrent neural networks (RNNs) capture time-varying dependencies via feedback weights that process information related to previous time steps \cite{Mandic2001}, that is 
\begin{equation}    \label{eq:RNNUnitht}
    \mathbf{h}_t = \sigma (\mathbf{W}^{hh} \mathbf{h}_{t-1} + \mathbf{W}^{xh} \mathbf{x}_t + \textbf{b}^h)
\end{equation} 
where the hidden state at time $t$, $\mathbf{h}_t \in \mathbb{R}^M$, is generated as a function of: (i) the previous hidden state $\mathbf{h}_{t-1} \in \mathbb{R}^M$, (ii) the input $\mathbf{x}_t \in \mathbb{R}^P$, (iii) the \textit{feedback} weight matrix $\mathbf{W}^{hh} \in \mathbb{R}^{M \times M}$, (iv) the \textit{input} weight matrix $\mathbf{W}^{xh} \in \mathbb{R}^{M \times P}$, and (v) a bias vector $\mathbf{b}^h \in \mathbb{R}^M$; $\sigma(\cdot)$ is an element-wise activation function. 

Similar to the application of TT to fully-connected neural networks, the folding of $\textbf{h}_t \in \mathbb{R}^M$, $\textbf{x}_t \in \mathbb{R}^P$, and $\textbf{b} \in \mathbb{R}^M$ into respective order-$N$ tensors, $\ten{H}_t \in \mathbb{R}^{J_1 \times J_2 \times \cdots \times J_N}$, $\ten{X}_t \in \mathbb{R}^{I_1 \times I_2 \times \cdots \times I_N}$, and $\ten{B} \in \mathbb{R}^{J_1 \times J_2 \times \cdots \times J_N}$, allows us to express $\mathbf{W}^{xh} \in \mathbb{R}^{M \times P}$ in the TT-format as in (\ref{eq:ttformat}). This results in the TT-RNN model \cite{Yang2017}, which extends the notion of tensorized neural networks to sequence modelling.

\begin{remark}
    \normalfont The TT-RNN compresses the parameter complexity for $\mathbf{W}^{xh}$ from an exponential $\prod_{n=1}^{N} I_n J_n = PM$ to a linear $\sum_{n=1}^N I_n J_n R_{n-1} R_n$ in the dimensions $I_n$ and $J_n$, which is highly efficient for small mode sizes and TT-ranks. This improves the storage complexity, computational complexity, and training time \cite{Yang2017}.
\end{remark}

%
%

\begin{table}[t!]
	\caption{Multi-way financial data selected from Bloomberg}
	\footnotesize
	\begin{center}
		\begin{tabular}{||c c c||} 
			\hline
			Asset Class & Symbol & Description \\ [0.5ex] 
			\hline\hline
			Equities (Index) & SPX & US stock index \\
			\hline
			Equities (Index) & MXCA & Canadian stock index \\
			\hline
			Equities (Index) & UKX & UK stock index \\
			\hline
			Equities (Index) & FTSEMIB & Italian stock index \\
			\hline
			Equities (Index) & SHSZ300 & Chinese stock index \\
			\hline
			Equities (Index) & NKY & Japanese stock index \\
			
			\hline
			Currencies (FX Pair) & CHFUSD & Swiss Frank \\
			\hline
			Currencies (FX Pair) & CADUSD & Canadian Dollar \\
			\hline
			Currencies (FX Pair) & GBPUSD & British Pound  \\
			\hline
			Currencies (FX Pair) & EURUSD & Euro  \\
			\hline
			Currencies (FX Pair) & CNYUSD & Chinese Yuan \\
			\hline
			Currencies (FX Pair) & JPYUSD & Japanese Yen  \\
			
			\hline
			Commodities (Futures) & GC1 & Gold \\
			\hline
			Commodities (Futures) & HG1 & Copper \\
			\hline
			Commodities (Futures) & CL1 & Crude Oil  \\
			\hline
			Commodities (Futures) & NG1 & Natural Gas \\
			\hline
			Commodities (Futures) & S1 & Soybeans \\
			\hline
			Commodities (Futures) & C1 & Corn \\
			
			\hline
			Fixed Income (Index) & USGG10YR & US 10 Yrs Yield \\
			\hline
			Fixed Income (Index) & GCAN10YR & Canadian 10 Yrs Yield  \\
			\hline
			Fixed Income (Index) & GUKG10 & UK 10 Yrs Yield \\
			\hline
			Fixed Income (Index) & GBTPGR10 & Italian 10 Yrs Yield \\
			\hline
			Fixed Income (Index) & GCNY10YR & Chinese 10 Yrs Yield \\
			\hline
			Fixed Income (Index) & GJGB10 & Japanese 10 Yrs Yield \\
			\hline
		\end{tabular}
		\label{tab:FDataSelection}
	\end{center}
\end{table}

\section{TT-RNN for Financial Forecasting} \label{sec:simulation}

\subsection{Input Data Structure}\label{sec:features}

To provide a comprehensive and multi-modal scheme to the task of financial forecasting of JPYUSD, we considered 24 financial signals with multi-way structure from Bloomberg, as shown in Table \ref{tab:FDataSelection}. The considered financial signals span four different asset classes: equities, currencies, commodities, and fixed income, with each class comprising of six financial components. The input feature vector at time instant, $t$, for $i$-th financial signal, $\mathbf{x}_t^i \in \mathbb{R}^{20}$, consists of $20$ features:
\begin{enumerate}
	\item Daily log-differences, $r_t = \log ({p_t}) - \log ({p_{t-1}})$, where $p_t$ is the daily closing value (\textbf{$\mathbf{1}$ feature});
	\item Rolling statistics of log-differences, such as the average, standard deviation, skewness, and kurtosis over $5, 10, 22$ time-steps, corresponding to $1$ week, $2$ weeks, and $1$ month of trading data ($\mathbf{12}$ \textbf{features});
	\item Relative price min-max $\frac{p_t - \min_{t \in T}(p_t)}{\max_{t \in T}(p_t) - \min_{t \in T}(p_t) }$, where $T = \{t-L+1, \dots, t-1, t\}$ and $L$ is a window of $5$, $10$, and $22$ time-steps  ($\mathbf{3}$ \textbf{features});
	\item Relative high-low-close, $\frac{p_t - l_t}{h_t - l_t}$, where $l_t$ and $h_t$ are the \textit{low} and \textit{high} price for day $t$ ($\mathbf{1}$ \textbf{feature});
	\item High-low spread, $\frac{h_t - l_t}{l_t}$ ($\mathbf{1}$ \textbf{feature});
	\item Units traded, i.e. the volume, at time $t$  ($\mathbf{1}$ \textbf{feature});
	\item Units of unsettled future contracts at time $t$ ($\mathbf{1}$ \textbf{feature}).
\end{enumerate}

\begin{figure}[t!]
	\centering
	\vspace{-3mm}
    \includegraphics[width=1\columnwidth]{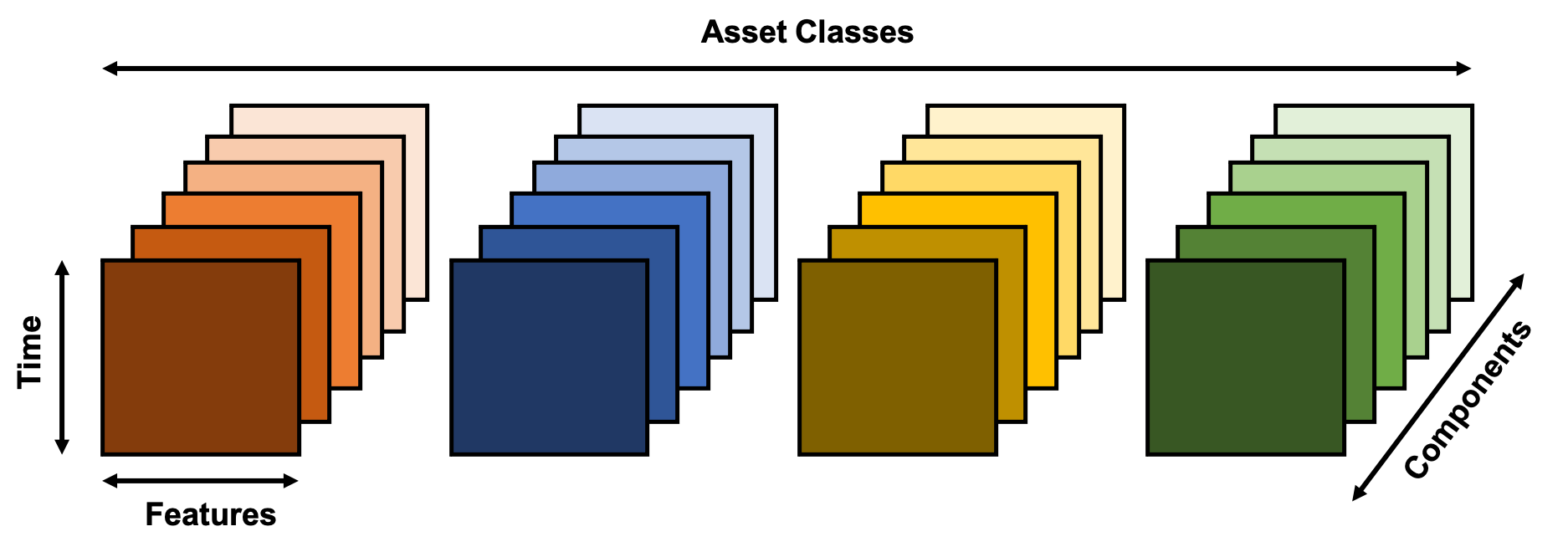}
	\vspace{-5mm}
	\caption{Financial data represented as a multi-modal tensor.}
	\vspace{-2mm}
	\label{fig:mm_finance}
\end{figure}

\vspace{2mm}
For every time-step, $t$, all 24 feature vectors $\mathbf{x}_t^i \in \mathbb{R}^{20}$ were arranged into a $3$-rd order tensor, $\ten{Z}_t \in \mathbb{R}^{20 \times 6 \times 4}$ (features $\times$ components $\times$ asset classes), which was further reshaped into a $5$-th order tensor, $\ten{X}_t \in \mathbb{R}^{2 \times 2 \times 5 \times 6 \times 4}$, such that the first three modes correspond to the $20$ features, while the last two correspond to the asset class components and asset classes. An intuitive illustration portraying the multi-modal nature of financial data in tensor form is presented in Figure \ref{fig:mm_finance}.

\begin{remark}
\normalfont The choice of a $5$-th order tensor, $\ten{X}_t$, instead of the original $3$-rd order tensor, $\ten{Z}_t$, improves the compression due to smaller mode sizes. This comes without loss in physical interpretability, as the modes corresponding to features, components, and asset classes are not mixed.
\end{remark}


\subsection{TT-RNN Training Settings} \label{sec:TTRNN_Training}

We considered the financial data in Table \ref{tab:FDataSelection} in the period from May $2006$ to May $2019$. The first $90\%$ of data was used for training, and the remaining 10\% for testing. Each feature was centred about its mean and normalized by its standard deviation. We employed a neural network consisting of one recurrent layer with 1024 units and \textit{tanh} activation, and one fully connected layer with 3 units and \textit{softmax} activation. The output labels $\{+1, 0, -1\}$ were assigned to the Japanese Yen time-series, according to whether the price at time step $(t+1)$ increased $(+1)$ or decreased $(-1)$ with respect to time step $t$; the value $0$ was assigned if $|r_t|\leq 10^{-4}$ to avoid unnecessary transaction costs in trading small market movements. Each input sample consisted of $10$ time-steps of $\mathcalbf{X}_t$. 

\begin{figure}[h]
	\centering
	\includegraphics[width=1\columnwidth]{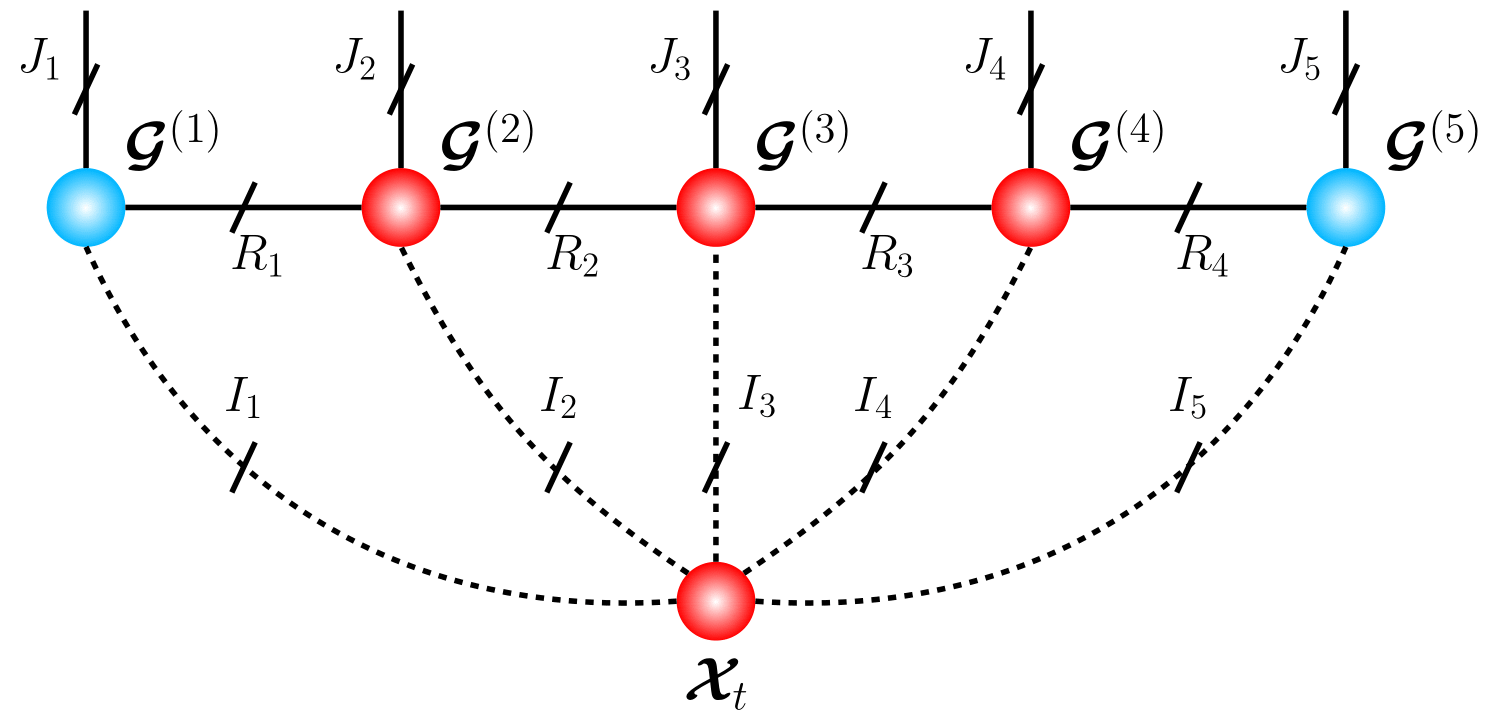}
	\caption{Modal connections between every mode of the input data tensor, $\ten{X}_t$, and the TT-cores, $\ten{G}_n$, of the tensorized weight matrix $\mathbf{W}^{xh}$ in TT-format. \vspace{-2mm}}
	\label{fig:basesview}
\end{figure}

We employed the TT-RNN model in the recurrent layer to compress the weight matrix $\mathbf{W}^{xh} \in \mathbb{R}^{M \times P}$ by reshaping and expressing it in the TT-format, in accordance to the dimensions of the input tensors $\ten{X}_t$, such that $P=2 \times 2 \times 5 \times 6 \times 4$ and $M=  4\times 4 \times 4\times 4 \times 4$. An illustration of the interaction between $\ten{X}_t$ and the tensorized $\mathbf{W}^{xh}$ is shown in Fig. \ref{fig:basesview} in Tensor Network form. 

The so established TT-RNN model was trained using stochastic gradient descent with a learning rate of $10^{-5}$ over 20 epochs, with a batch size of 66, and using the categorical cross-entropy loss function. The TT-ranks $(R_1, R_2, R_3, R_4)$ for the TT-RNN model were empirically found to be optimal for rank $R_i = 6$, $i=1, \dots, 4$. It was also found that higher values of $R_i$ would result in over-fitting, and lower values in under-fitting. In turn, this suggests that the TT-ranks may be used for regularization in addition to compression. 

\begin{remark}
	\normalfont Fig. \ref{fig:basesview} indicates that by keeping track of the variation of each core during training allows the TT-RNN model to: (i) reveal information about the relative importance of each mode, and (ii) achieve network compression and regularization \cite{Calvi2019_2}.
\end{remark}

\subsection{Experimental Results}

\begin{figure}[h!]
	\centering
	\includegraphics[width=1\columnwidth]{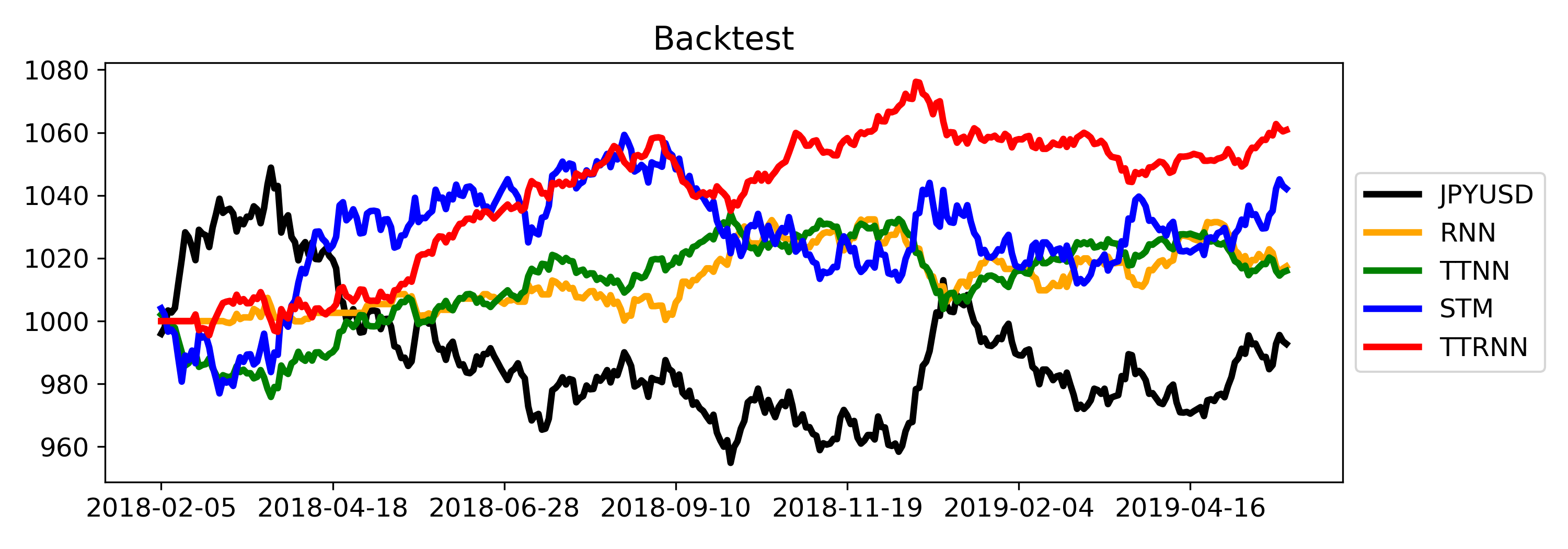}
	\caption{Cumulative profits generated by the models.}
	\label{fig:Backtest}
\end{figure}

For the given task, we compared the performance of the proposed TT-RNN against: (i) its uncompressed RNN counterpart, (ii) a TT fully-connected neural network (TTNN) \cite{Novikov2015}, and (iii) a Support Tensor Machine (STM) \cite{calvi2019support}. 

Fig. \ref{fig:Backtest} shows the simulated cumulative profits generated by trading according to model predictions over the test period. The trades were updated daily and sized proportionally to the probability estimates of the model. A buy-and-hold strategy of the underlying asset JPYUSD is also shown in the figure.

To evaluate the performance of each model, we considered: (i) the annualized Sharpe ratio of the profits generated in the test-period (defined as $\sqrt{252}\frac{u_r}{\sigma_r}$, where $u_r$ is the average daily returns and $\sigma_r$ is the standard deviation of daily returns), (ii) total returns generated, and (iii) the directional accuracy of the predictions. The results are summarized in Table \ref{tab:metrics}. 

By unifying the regularization property of tensors and the sequence modelling ability of RNNs, the proposed TT-RNN out-performed its uncompressed RNN counterpart and other general purpose tensor models such as TTNN and STM, obtaining the best experimental results across all metrics.

\begin{table}[h!]
\caption{Performance comparison of the considered models.}
\begin{center}
\begin{tabular}{|l|lll|}
\hline
               & Sharpe         & Total Return    & Accuracy         \\ \hline
JPYUSD         & -0.014         & -0.73\%         & N.A.             \\
RNN            & 0.491          & 1.74\%          & 36.05\%          \\
TTNN           & 0.372          & 1.60\%          & 51.06\%          \\
STM            & 0.520          & 4.21\%          & 51.67\%          \\
\textbf{TTRNN} & \textbf{1.548} & \textbf{6.08\%} & \textbf{52.04\%} \\ \hline
\end{tabular}
\vspace{-3mm}
\label{tab:metrics}
\end{center}
\end{table}

\begin{remark}
	\normalfont
	Due to the low signal-to-noise ratio of financial data, a classification accuracy of above 50\% is considered significant for financial applications.
\end{remark}

\subsection{Interpreting Relative Modal Importance}

\begin{figure}[h!]
	\centering
    \includegraphics[width=1\columnwidth]{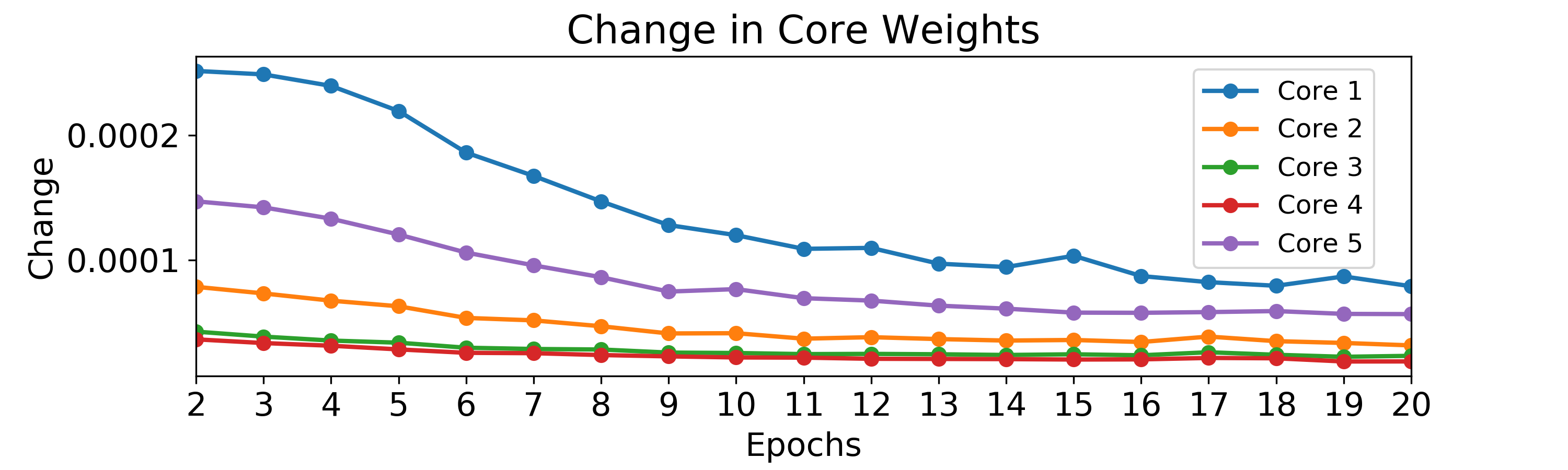}
	\caption{Normalized change of TT-core weights in TT-RNN.}
	\label{fig:TTRMdlWCurve2}
\end{figure}

In addition to the compression and regularization power of the TT decomposition, the multi-modal structure inherent to the TT allowed us to interpret the relative modal importance of the data, by indicating which data modality experienced largest changes during training. This is illustrated in Fig. \ref{fig:TTRMdlWCurve2}, which shows the normalized TT-core change over the training epochs, defined as
\begin{equation}
	\frac{||\Delta \ten{G}^e_{n}||_F^2}{I_n J_n R_{n-1} R_n}
\end{equation}
where $\Delta \ten{G}^e_{n} = \ten{G}^e_{n} - \ten{G}^{e-1}_{n}$ and $\ten{G}^e_{n}$ is the value of $\ten{G}_{n}$ at epoch $e=2,\ldots,20$. 

The cores $\ten{G}_{1}$ and $\ten{G}_{5}$ exhibited greatest changes, suggesting that the hand-crafted features contained in mode 1 (see Section \ref{sec:features}), and the asset classes contained in mode 5 (see Table \ref{tab:FDataSelection}), account for most of the predictive power of TT-RNN for our problem. Also, the greater change in $\ten{G}_{5}$ over $\ten{G}_{4}$ suggests greater importance of inter-class dependencies (mode-5) over intra-class financial signals (mode-4). This insight is only possible with the weights represented in the TT format, which associates a TT core to each of the data modality separately, hence bringing the benefits of interpretability in addition to the compression and regularization properties of TTD.



\section{Conclusion}

Traditional currency forecasting models are typically based on intra-class variables only, despite the wide availability of other strongly or weakly related asset classes. To address this issue, we investigated the ability of Tensor-Train Recurrent Neural Networks (TT-RNNs) to perform financial forecasting of currencies from multi-class data. This has been achieved based on multi-way financial data from four asset classes: equities, currencies, commodities, and fixed income. The model has been trained on tensorized data from May 2006 to February 2018, and tested on the period from February 2018 to May 2019. Simulation results have verified the desirable compression and regularization properties of the TT-RNN, which outperformed its uncompressed RNN counterpart and other tensor based forecasting methods. Moreover, through the gradient inspection of individual TT cores, TT-RNN has been demonstrated to allow for physical interpretability of relative modal importance in a financial context, thus mitigating the well-known black-box issue inherent to neural networks. 

\bibliographystyle{IEEEtran.bst}
\bibliography{references.bib}

\begin{thebibliography}{10}
\providecommand{\url}[1]{#1}
\csname url@samestyle\endcsname
\providecommand{\newblock}{\relax}
\providecommand{\bibinfo}[2]{#2}
\providecommand{\BIBentrySTDinterwordspacing}{\spaceskip=0pt\relax}
\providecommand{\BIBentryALTinterwordstretchfactor}{4}
\providecommand{\BIBentryALTinterwordspacing}{\spaceskip=\fontdimen2\font plus
\BIBentryALTinterwordstretchfactor\fontdimen3\font minus
  \fontdimen4\font\relax}
\providecommand{\BIBforeignlanguage}[2]{{%
\expandafter\ifx\csname l@#1\endcsname\relax
\typeout{** WARNING: IEEEtran.bst: No hyphenation pattern has been}%
\typeout{** loaded for the language `#1'. Using the pattern for}%
\typeout{** the default language instead.}%
\else
\language=\csname l@#1\endcsname
\fi
#2}}
\providecommand{\BIBdecl}{\relax}
\BIBdecl

\bibitem{Mandic2001}
D.~P. Mandic and J.~Chambers, \emph{Recurrent neural networks for prediction:
  Learning algorithms, architectures and stability}.\hskip 1em plus 0.5em minus
  0.4em\relax John Wiley, Inc., 2001.

\bibitem{He2017}
Z.~He, S.~Gao, L.~Xiao, D.~Liu, H.~He, and D.~Barber, ``Wider and deeper,
  cheaper and faster: {T}ensorized {LSTMS} for sequence learning,''
  \emph{Proceedings of {A}dvances in {N}eural {I}nformation {P}rocessing
  {S}ystems (NIPS)}, pp. 1--11, 2017.

\bibitem{khrulkov2017expressive}
V.~Khrulkov, A.~Novikov, and I.~Oseledets, ``Expressive power of recurrent
  neural networks,'' in \emph{International Conference on Learning
  Representations}, 2018.

\bibitem{Mandic2015}
A.~Cichocki, D.~P. Mandic, A.~H. Phan, C.~F. Caiafa, G.~Zhou, Q.~Zhao, and
  L.~D. Lathauwer, ``{Tensor decompositions for signal processing
  applications},'' \emph{IEEE Signal Processing Magazine}, vol.~32, no.~2, pp.
  145--163, 2015.

\bibitem{Kolda2009}
T.~Kolda and B.~Bader, ``{Tensor decompositions and applications},'' \emph{SIAM
  Review}, vol.~51, no.~3, pp. 455--500, 2009.

\bibitem{sidiropoulos2017tensor}
N.~D. Sidiropoulos, L.~De~Lathauwer, X.~Fu, K.~Huang, E.~E. Papalexakis, and
  C.~Faloutsos, ``Tensor decomposition for signal processing and machine
  learning,'' \emph{IEEE Transactions on Signal Processing}, vol.~65, no.~13,
  pp. 3551--3582, 2017.

\bibitem{Mandic2016_2}
A.~Cichocki, N.~Lee, I.~Oseledets, A.~H. Phan, Q.~Zhao, and D.~P. Mandic,
  ``Tensor networks for dimensionality reduction and large-scale optimization.
  {P}art 1: {L}ow-rank tensor decompositions,'' \emph{Foundations and Trends in
  Machine Learning}, vol.~9, no. 4-5, pp. 249--429, 2016.

\bibitem{Mandic2017}
A.~Cichocki, A.-H. Phan, Q.~Zhao, N.~Lee, I.~Oseledets, M.~Sugiyama, and D.~P.
  Mandic, ``Tensor networks for dimensionality reduction and large-scale
  optimization. {P}art 2: {A}pplications and future perspectives,''
  \emph{Foundations and Trends in Machine Learning}, vol.~9, no.~6, pp.
  431--673, 2017.

\bibitem{Kim2015}
Y.~D. Kim, E.~Park, S.~Yoo, T.~Choi, L.~Yang, and D.~Shin, ``Compression of
  deep convolutional neural networks for fast and low power mobile
  applications,'' \emph{arXiv:1511.06530}, 2015.

\bibitem{Zhong2019}
Z.~Zhong, F.~Wei, Z.~Lin, and C.~Zhang, ``{ADA-Tucker: Compressing deep neural
  networks via adaptive dimension adjustment {T}ucker decomposition},''
  \emph{Neural Networks}, vol. 110, pp. 104--115, 2019.

\bibitem{Calvi2019_2}
G.~G. Calvi, A.~Moniri, M.~Mahfouz, Z.~Yu, Q.~Zhao, and D.~P. Mandic, ``Tucker
  tensor layer in fully connected neural networks,'' \emph{arXiv:1903.06133},
  2019.

\bibitem{lebedev2014speeding}
V.~Lebedev, Y.~Ganin, M.~Rakhuba, I.~Oseledets, and V.~Lempitsky, ``Speeding-up
  convolutional neural networks using fine-tuned cp-decomposition,''
  \emph{arXiv preprint arXiv:1412.6553}, 2014.

\bibitem{Novikov2015}
A.~Novikov, D.~Podoprikhin, A.~Osokin, and D.~P. Vetrov, ``Tensorizing neural
  networks,'' in \emph{Proceedings of the Advances in Neural Information
  Processing Systems (NIPS)}, 2015, pp. 442--450.

\bibitem{Oseledets2011}
I.~V. Oseledets, ``Tensor-train decomposition,'' \emph{SIAM Journal on
  Scientific Computing}, vol.~33, no.~5, pp. 2295--2317, 2011.

\bibitem{Yang2017}
Y.~Yang, D.~Krompass, and V.~Tresp, ``Tensor-train recurrent neural networks
  for video classification,'' \emph{in proceedings of the 34th International
  Conference on Machine Learning-Volume 70}, pp. 3891--3900, 2017.

\bibitem{yu2017}
R.~Yu, S.~Zheng, A.~Anandkumar, and Y.~Yue, ``Long-term forecasting using
  tensor-train {RNN}s,'' \emph{Arxiv}, 2017.

\bibitem{tjandra2017compressing}
A.~Tjandra, S.~Sakti, and S.~Nakamura, ``Compressing recurrent neural network
  with tensor train,'' in \emph{Proceedings of 2017 International Joint
  Conference on Neural Networks (IJCNN)}.\hskip 1em plus 0.5em minus
  0.4em\relax IEEE, 2017, pp. 4451--4458.

\bibitem{de2020machine}
M.~L. De~Prado, \emph{Machine Learning for Asset Managers}.\hskip 1em plus
  0.5em minus 0.4em\relax Cambridge University Press, 2020.

\bibitem{de2018advances}
M.~L. de~Prado, \emph{Advances in Financial Machine Learning}.\hskip 1em plus
  0.5em minus 0.4em\relax John Wiley \& Sons, 2018.

\bibitem{dees2019analysing}
B.~S. Dees, ``Analysing global fixed income markets with tensors,'' \emph{arXiv
  preprint arXiv:1908.02101}, 2019.

\bibitem{tsantekidis2017using}
A.~Tsantekidis, N.~Passalis, A.~Tefas, J.~Kanniainen, M.~Gabbouj, and
  A.~Iosifidis, ``Using deep learning to detect price change indications in
  financial markets,'' in \emph{Proceedings of 25th European Signal Processing
  Conference (EUSIPCO)}.\hskip 1em plus 0.5em minus 0.4em\relax IEEE, 2017, pp.
  2511--2515.

\bibitem{tran2018temporal}
D.~T. Tran, A.~Iosifidis, J.~Kanniainen, and M.~Gabbouj, ``Temporal
  attention-augmented bilinear network for financial time-series data
  analysis,'' \emph{IEEE Transactions on Neural Networks and Learning Systems},
  vol.~30, no.~5, pp. 1407--1418, 2018.

\bibitem{tran2017tensor}
D.~T. Tran, M.~Magris, J.~Kanniainen, M.~Gabbouj, and A.~Iosifidis, ``Tensor
  representation in high-frequency financial data for price change
  prediction,'' in \emph{Proceedings of IEEE Symposium Series on Computational
  Intelligence (SSCI)}, 2017, pp. 1--7.

\bibitem{dixon2018sequence}
M.~Dixon, ``Sequence classification of the limit order book using recurrent
  neural networks,'' \emph{Journal of computational science}, vol.~24, pp.
  277--286, 2018.

\bibitem{tsantekidis2020using}
A.~Tsantekidis, N.~Passalis, A.~Tefas, J.~Kanniainen, M.~Gabbouj, and
  A.~Iosifidis, ``Using deep learning for price prediction by exploiting
  stationary limit order book features,'' \emph{Applied Soft Computing},
  vol.~93, p. 106401, 2020.

\bibitem{gal2015theoretically}
Y.~Gal and Z.~Ghahramani, ``A theoretically grounded application of dropout in
  recurrent neural networks,'' \emph{arXiv preprint arXiv:1512.05287}, 2015.

\bibitem{dieng2018noisin}
A.~B. Dieng, R.~Ranganath, J.~Altosaar, and D.~Blei, ``Noisin: Unbiased
  regularization for recurrent neural networks,'' in \emph{Proceedings of
  International Conference on Machine Learning}.\hskip 1em plus 0.5em minus
  0.4em\relax PMLR, 2018, pp. 1252--1261.

\bibitem{Dolgov2014}
S.~Dolgov and D.~Savostyanov, ``{Alternating minimal energy methods for linear
  systems in higher dimensions},'' \emph{SIAM Journal on Scientific Computing},
  vol.~36, no.~5, pp. A2248--A2271, 2014.

\bibitem{Cichocki2014}
A.~{Cichocki}, ``Era of big data processing: A new approach via tensor networks
  and tensor decompositions,'' \emph{ArXiv e-prints}, Mar. 2014.

\bibitem{Oseledets2009}
I.~V. Oseledets and E.~E. Tyrtyshnikov, ``Breaking the curse of dimensionality,
  or how to use {SVD} in many dimensions,'' \emph{SIAM Journal on Scientific
  Computing}, vol.~31, no.~5, pp. 3744--3759, 2009.

\bibitem{calvi2019support}
G.~G. Calvi, V.~Lucic, and D.~P. Mandic, ``Support tensor machine for financial
  forecasting,'' in \emph{Proceedings of IEEE International Conference on
  Acoustics, Speech and Signal Processing (ICASSP)}, 2019, pp. 8152--8156.

\end{thebibliography}

\end{document}